\documentclass[11pt,a4paper]{article}

\usepackage{microtype}
\usepackage{graphicx}
\usepackage{subfigure}
\usepackage{booktabs} 
\usepackage{amsmath}
\usepackage{amssymb}
\usepackage{hyperref}

\newcommand{\dk}{d_{k}}
\newcommand{\dv}{d_{v}}

\newcommand{\policy}[1][w]{{\pi({#1}|{\phi})}}

\newcommand{\Att}[1]{{\textrm{Att}_{#1}}}
\newcommand{\FF}{{\textrm{FF}}}
\newcommand{\Dense}[1][d]{{\textrm{Dense}_{#1}}}
\newcommand{\LayerNorm}{{\textrm{LN}}}
\newcommand{\LN}{{\textrm{LN}}}
\newcommand{\Conn}{{\Psi}}
\newcommand{\Act}{{\textrm{Activation}}}
\newcommand{\Softmax}{{\textrm{Softmax}}}
\newcommand{\Sim}{{\textrm{Sim}}}

\newcommand{\Res}{{\textrm{Res}}}
\newcommand{\Concat}{{\textrm{Concat}}}
\newcommand{\Combine}{{\Omega}}

\newcommand{\Lcost}[1][w]{{L_c(#1)}}
\newcommand{\Lorig}[1][w,\theta]{{L_o({#1})}}
\newcommand{\Ltotal}[1][w,\theta]{{L({#1})}}
\newcommand{\Lsample}[1][\theta,\phi]{{L'({#1})}}

\DeclareMathOperator*{\E}{\mathbb{E}}

\DeclareMathOperator*{\argmin}{arg\,min}

\usepackage[hyperref]{emnlp2020}
\usepackage{times}
\usepackage{latexsym}

\usepackage{microtype}

\aclfinalcopy 

\title{Finding Fast Transformers: One-Shot Neural Architecture Search by Component Composition}

\author{Henry Tsai, Jayden Ooi, Chun-Sung Ferng, Hyung Won Chung, Jason Riesa\\
Google Research\\
\texttt{\{henrytsai, jayden, csferng, hwchung, riesa\}@google.com}}

\date{}

\begin{document}
\maketitle
\begin{abstract}
Transformer-based models have achieved state-of-the-art results in many tasks in natural language processing. However, such models are usually slow at inference time, making deployment difficult. In this paper, we develop an efficient algorithm to search for fast models while maintaining model quality. We describe a novel approach to decompose the Transformer architecture into smaller components, and propose a sampling-based one-shot architecture search method to find an optimal model for inference. The model search process is more efficient than alternatives, adding only a small overhead to training time. By applying our methods to BERT-base architectures, we achieve 10\% to 30\% speedup for pre-trained  BERT and 70\% speedup on top of a previous state-of-the-art distilled BERT model on Cloud TPU-v2 with a generally  acceptable drop in performance.
\end{abstract}

\section{Introduction}

Deep residual models like the Transformer \cite{NIPS2017_7181} have achieved state-of-the-art results on many tasks.  However, the most accurate models are usually slow at inference time, making real-world deployment prohibitive for many applications.

In this paper we describe a novel approach to finding the optimal architecture for Transformer networks, optimizing for inference time and maintaining accuracy. The final model can be trained from scratch or used in conjunction with techniques like \textit{distillation} \cite{hinton2015}.

We propose a component-wise network selection approach for Transformer-based networks. We use BERT \cite{DBLP:journals/corr/abs-1810-04805} as a running example to show how to construct components based on the need for high-speed models and formulate an objective that is directly relevant to inference speed. Then, we propose a simple sampling-based model selection algorithm that automatically selects hyperparameters in one-shot. Our contributions are as follows:
\begin{itemize}
\setlength\itemsep{0.001em}
   \item We propose a novel re-parameterization of the Transformer which enables us to search the depth and width of the model at the same time.
    \item We design an objective to integrate external computation profiling information and formulate the objective to directly optimize for high-speed models.
    \item We propose a sampling-based algorithm for one-shot model selection that yields state-of-the-art results while adding little memory overhead and small (1.4x) training time overhead. 
\end{itemize}
We evaluate our methods on a wide variety of tasks and show how fast model architectures can vary for different tasks.

\section{Background and Related Work}

\subsection{Smaller Models}
Smaller models do not always mean faster models: the total number of parameters does not necessarily correlate with the amount of computation needed. For example, the softmax operation itself is not assigned parameters but can be expensive when the output dimension is large; the embedding dimension can be large, but the embedding-lookup operation can be light-weight if the operation is optimized. 

A significant amount of previous work is concerned with model-size minimization. There are several typical approaches, and we briefly examine how each contributes to model inference speedup. Most work described below is orthogonal to this work, and as a result can be applied on top of the methods we introduce in this paper.

\paragraph{Quantization} Previous work has shown that instead of using 32-bit floats to store the weights, models can be quantized to 8-bit or even 4-bit floating point numbers of integers without much accuracy loss~\cite{1910.06188, 2004.07320}. By doing that, one can easily obtain a model that is 4x smaller or more. Quantized models may run faster on hardware that support quantized arithmetic \cite{Jacob_2018_CVPR}.

\paragraph{Sparsity} Model sparsity, or zeroing-out model parameters, is another common minimization approach \cite{1710.01878}; zero-weights typically do not need to be stored. However, keeping track of sparse matrices may add additional computational overhead. Usually speedups are only observed if the model is very sparse (e.g. 90\% sparsity)  and is running on hardware that supports sparse operations well.\footnote{For example, Intel's sparse matrix kernel library.} 
Alternatively, structured sparsity \cite{DBLP:journals/corr/abs-1711-06798} which adds constraints to introduce sparsity on each tensor row may help achieve better speedup since row pruning reduces the dimensionality of the tensors. One problem with such methods is that it is limited by the existing network architecture.

\paragraph{Parameter Sharing} We can obtain smaller models by sharing parameters across layers \cite{Lan2020ALBERT}. However, resulting models are typically not much faster because computation is not shared. In addition, a larger architecture may be required, affecting the computation graph, in order to compete in tasks with models that have no parameter sharing.

\subsection{Models with Fewer FLOPs}
There are many existing works that try to minimize FLOPs\footnote{floating-point operations per second} \cite{DBLP:journals/corr/abs-1711-06798}. However, fewer FLOPs do not always mean faster speed. A model can have increased FLOPs but still run faster at inference time because it uses the computational hardware more effectively. 

\subsection{One-Shot Neural Architecture Search}

There are many ways to speed up the traditionally slow neural architecture search process \cite{JMLR:v20:18-598}. One recent focus is one-shot search. In such methods, only one model is trained, and the final model is just a sub-network of the one-shot model. 

We can think of a  neural network as a directed acyclic graph with different functions on the edges of the graph. Given $n$ nodes and $k$ sub-network candidates for each edge, including dropout of the edge connection itself, we can search all $O(n^2k)$ combinations to find the optimal model. Doing one-shot search means that one needs to search all sub-network candidates at the same time. As a result, such methods usually use a lot of memory to store the weights of all candidates, and applying them on state-of-the-art large networks can be difficult. 

There are two categories of algorithms to search the combinations: direct pruning methods \cite{DBLP:journals/corr/abs-1711-06798} and sampling-based methods \cite{ DBLP:journals/corr/ShazeerMMDLHD17}, including reinforcement-learning  methods \cite{xie2018snas}. Regularizers can be added to the network to make the final model having certain properties. Many works are proposed in both categories, but there are few works that compare them directly. To our knowledge, this is the first work to compare the two in a controlled setting.

\section{Architecture Search Space}
The standard BERT model architecture consists of a series of Transformer blocks, each containing a multi-headed attention followed by a $2$-layer position-wise feedforward block. There is a residual connection around each attention and feedforward block, after which the output is passed through layer normalization.

Based on profiling results, we have identified the following network components and hyperparameters as having substantial impact on inference efficiency, and design our search space around them:
\begin{itemize}
\setlength\itemsep{0.001em}
    \item Attention query key and value dimensions
    \item Width and depth of feedforward layers
    \item Number of attention heads
    \item Layer normalization mean computation
\end{itemize}

We do not require each Transformer block to share the same structure, as required by some previous work \cite{DBLP:journals/corr/abs-1901-11117}.
Unlike the standard Transformer, we explore dimensions for feedforward and attention key-value,
independent of the hidden layer size and the number of attention heads.

To manage the search space due to combinatorial explosion, we formulate the model as a sequence of function compositions and represent choices of the hyperparameters as searching the composition of smaller components to avoid searching all possible sizes. By allowing components to share weights as model architecture changes, we can search many architectures without retraining and with little memory overhead.

\subsection{Network Component}
\label{section:size-selection}
This section defines the key network components and the corresponding hyperparameters that are crucial for constructing our search space. In order to search for different component dimensions, we derive the decomposition for each component that enables the search algorithms to optimize for different dimensions later.

We use \emph{component} to refer to any learnable function or sub-network.
To simplify the notations, we only describe the component type and omit the different learnable parameters.
Let $X$ denotes the input with dimension $[\ell, h]$, corresponding to the sequence length and hidden layer size respectively.


\paragraph{Feedforward Network}
The position-wise feedforward network in a Transformer block has $2$ fully-connected layers. Both input and output size are fixed to $h$, while the intermediate dimension $d$ is flexible. Let $\FF_d$ denote a $2$-layer with intermediate dimension $d$. We have the decomposition
\begin{align*}
\begin{split}
\FF_d(X) =& \hspace{2pt} \Dense[h](\Act(\Dense[d](X))) \\
=& \sum_{i=1}^m \Dense[h](\Act(\Dense[\frac{d}{m}](X))) \\
=& \sum_{i=1}^m \FF_{\frac{d}{m}}(X),
\end{split}
\end{align*}
which is a summation of $m$ feedforward networks of size $d/m$ each.
 
\paragraph{Query-Key Similarity} 
Query-key similarity is the core operation in the attention mechanism. Given the component with key dimension $\dk$, we can decompose it into
\begin{align*}
\begin{split}
\Sim_{\dk}(X)
&= \Dense[\dk](X)(\Dense[\dk](X))^T \\
&= \sum_{i=1}^m \Sim_{\frac{\dk}{m}}(X),
\end{split}
\end{align*}
where each of the $m$ parts just have a smaller key dimension of $\dk/m$.

\paragraph{Multi-Head Attention}
Let $\Att{a, \dk, \dv}$  denote an $a$-head self-attention with key and value dimension of $\dk$, $\dv$ respectively.
As multi-head attention is the concatenation of all heads' output followed by a linear projection, we can naturally divide it to a summation of single-head attentions.
\begin{align*}
\begin{split}
\Att{a, \dk, \dv}(X) =& \Dense[h](\Concat(\textrm{Head}_1, \ldots, \textrm{Head}_a)) \\
=& \sum_{i=1}^a \Att{1, \dk, \dv}(X)
\end{split}
\end{align*}

\paragraph{Single-Head Attention}
Similarly, the \emph{attention value} computation of single-head attention can be broken into $m$ equal parts.

\begin{align*}
\begin{split}
&\Att{1, \dk, \dv}(X) \\
&= \Dense[h]\left(\Softmax\left(\frac{\Sim_{\dk}(X)}{\sqrt{\dk}}\right) \cdot \Dense[\dv](X)\right) \\
&= \sum_{i=1}^m \Att{1,\dk, \frac{\dv}{m}}(X)
\end{split}
\end{align*}

We have shown that all components mentioned above can be decomposed into summation of $m$ equal parts. Now, for each sub-component $f_i(X)$, define a corresponding binary variable $w_i$ of whether to keep that sub-component, so component output can be written as $\sum_i w_i f_i(X)$. The selection parameters can be optimized by the search algorithm, and setting any $w_i$ to $0$ effectively reduces the component dimension (e.g. $d$ for feedforward, and $a$, $\dk$, $\dv$ for attention).

\paragraph{Layer Normalization}
Some existing works have shown that zero-mean normalization in batch normalization is not needed \cite{2003.07845}.
We explore the same for layer normalization by replacing the mean with $\mu'(X) = w \cdot \mu(X)$ conditioned on a selection parameter $w$, giving
\begin{align*}
\LayerNorm(X) &= \frac{\alpha\cdot(X-\mu'(X))}{\sigma} + \beta,
\end{align*}
where $\sigma = \sqrt{(X-\mu')^2 / N}$.
The search algorithm can disable zero-mean normalization by assigning $w=0$.

\subsection{Architecture Connection}
Finding the best layer width and depth for a given network size is another different challenge.
The search space consists of exponentially many possible configurations that we need to be able to represent and optimize on.

To achieve that, imagine a sequence of $k$ identical components (e.g., \FF) to be assembled in a network. Each component $f_i$, except the last, can either be placed in the same layer as its successor (horizontal connection), or in a different layer (vertical connection).
Similarly, we define a \emph{connection parameter} $w_i \in \{0, 1\}$ to represent these two choices respectively.
Notice that the $2^{k-1}$ possible choices corresponds exactly with all possible layer configurations. This view provides a useful mean for constructing our search space.

We also need an accumulating mechanism that can combine output of all components $f_i(X)$ in the same layer. This can be implemented in a network by passing an accumulated memory $R$ as additional input across the components.

To be able to represent any residual networks, it's crucial to include residual connection in the search space.


\subsubsection{Connector Unit}
\begin{figure}[bt]
\centering
\includegraphics[width=8cm]{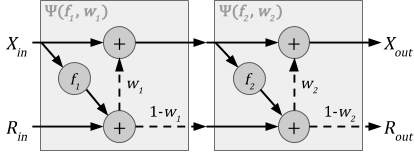}
\caption{$\Conn(f_2,w_2)\circ\Conn(f_1,w_1)$. The dashed arrows are multiplied by edge weights.
When $w_1=0$, $f_1$ and $f_2$ share the same input, giving a horizontal connection.
When $w_1=1$, $f_2$ takes $X+f_1(X)+R$ as input, thus $f_2$ is in the next layer and the connection is vertical.
}
\label{fig:vertical_vs_horizontal}
\end{figure}

We define \emph{connector unit} $\Conn$ as a higher-order function that takes a basic component $f$ and connection parameter $w$, and outputs the function
\begin{align*}
\begin{split}
\Conn&(f, w)(X, R) =\\
&(X + w \cdot(f(X) + R), (1-w) \cdot (f(X) + R)).
\end{split}
\end{align*}
Output of $\Conn(f, w)$ is a tuple to be fed to the next connector unit, as illustrated in Figure~\ref{fig:vertical_vs_horizontal}.


Input $R$ contains the cumulative output for current layer up until current component.
When $w=0$, $f(X)$ is added to $R$ to continue accumulating current layer's output, while $X$ is passed through unchanged.
When $w=1$, current layer is concluded by summing the input $X$, current output $f(X)$, and cumulative output $R$ together. A more detailed example can be found in Appendix~\ref{appendix:connector-example}.



In order to interoperate $\Conn(\cdot)$ with unary functions, we define $\Combine(X, R) = X+R$, and with a slight abuse of notation, let $\Conn(f,w)(X) = \Conn(f,w)(X, 0)$ when only a single input is given.

\subsubsection{Residual Connection}
As shown earlier, feedforward networks with dimension $d$ is equivalent to a summation of $m$ feedforward networks with dimension $d/m$. Using the connector unit defined above, a residual-added feedforward network can be expressed as $m$ horizontally connected networks.
\begin{align*}
\begin{split}
&\Res(\FF_d)(X) = X + \sum_{i=1}^m \FF_{\frac{d}{m}}(X) \\
& =\left(\Combine\circ\left(\mathop{\bigcirc}\limits_{i=1}^{m}\Conn\left(\FF_{\frac{d}{m}}, 0 \right)\right)\right)(X)
\end{split}
\end{align*}
Detailed derivation is provided in Appendix~\ref{appendix:res_to_conn}. Similarly we can write multi-head attention as horizontally connected single-head attentions.
\begin{align*}
\begin{split}
&\Res(\Att{a,\dk,\dv})(X) = X+\sum_{i=1}^a \Att{1, \dk, \dv}(X) \\
&= \left(\Combine\circ\left(\mathop{\bigcirc}\limits_{i=1}^{a}\Conn(\Att{1,\dk,\dv}, 0)\right)\right)(X)
\end{split}
\end{align*}

Thus a Transformer block can be rewritten as
\begin{align*}
\begin{split}
&\left(\LN\circ\Combine\circ
\left(\mathop{\bigcirc}\limits_{i=1}^{m}\Conn\left(\FF_{\frac{d}{m}}, w_i \right)\right)\circ\right. \\ 
&\left.\LN\circ\Combine\circ
\left(\mathop{\bigcirc}\limits_{j=1}^{a}\Conn(\Att{1, \dk, \dv}, w_j)\right)\right)(X),
\end{split}
\end{align*}
where $w_i=w_j=0$ for all $i$, $j$. One can choose a different $m$, or change any of $w_i$ and $w_j$ to get model architectures with various width and depth of the feedforward and attention networks for a given model size.

\subsection{Search Space Considerations}

We have established our architecture search space, parameterized by selection parameters that determines which sub-components to retain, and connection parameters that controls the connection orientations.
We will refer to them more generally as \emph{architecture parameters}, still denoted by $w$.
Let $\theta$ be other \emph{network parameters} that do not affect the architecture. Both $w$ and $\theta$ are jointly trained to produce a complete model.

The choice of $m$ to divide the sub-components into allows us to control the granularity of the search space. If $m$ is too small, each component may be too large that even dropping one hurts the model quality.
If $m$ is too large, besides expanding the search space, it can result in many small components that each require fewer FLOPS to compute, but may have worse device utilization overall.

We also exclude incompatible or known inefficient settings from the search space. For example, two heads with different $\dv$ in the same multi-head attention will limit parallelization of multi-head attention, and therefore not considered.

\section{One-Shot Search}
For each distinct component $f_i$, we run offline profiling on the target device to measure its computation cost $c_i$. 
Let $\Lcost$ be the total network cost, which depends on the costs $c = \{c_i\}$ and architecture parameters $w$, but not $\theta$.
Let $\Lorig$ denote the loss function in the original problem that is a function of both network weights and architecture. 
The goal is to find the optimal
$$w^*, \theta^* = \argmin_{w, \theta} \Lcost,$$
subjected to the constraint that $\Lorig$ is no worse than some baseline. To simplify computation, we relax the constraint optimization problem into minimizing 
$$\Ltotal = \Lorig + \lambda \Lcost,$$
where $\lambda$ is a tunable hyperparameter.

Notice that in our search space formulations, $w_i=0$ corresponds to either dropping the component or connecting horizontally, both of which don't add incremental cost to the network. Using this observation, the total cost can be approximated as
$$\Lcost = \sum_i w_i \cdot c_i,\;\text{where}\; w_i \in \{0, 1\}.$$
This approximation correlates well with the computation time from our experiments.

\subsection{Direct Optimization (DO)}
Optimizing the cost with integer constraints is intractable in general. In this method, we relax the constraints on $w$ and use $\Lcost = \sum_i |w_i| \cdot c_i$ in minimizing the total loss $\Ltotal$. This resembles $L_1$ norm regularization that encourages sparse solution, which is a desirable outcome.
After training, we prune components with $|w_i|$ below threshold ($10^{-6}$) to get a leaner model.

The optimization may yield some $w_i\notin\{0, 1\}$, which  make $\Lcost$ a less accurate estimate of the computation cost. Nevertheless, the final network is still valid with the following interpretation. For selection parameters, as they simply scale the components' output, that is equivalent to $w_i=1$ and re-scaling $\theta$ accordingly. If a connection parameter is not $0$ or $1$,
it represents a scaled residual connection to the next connector unit.

\subsection{Sampling Distribution Optimization (SDO)}
A disadvantage of DO is the inability to enforce $w_i \in \{0, 1\}$ and having many components with small weight could be a source of inefficiency. Besides, it is not possible to coordinate selection / connection decisions by incorporating more sophisticated dependencies between decisions.

Instead of learning $w$ itself, we learn a sampling distribution or \emph{policy} $\policy$ to sample $w$ for training, where $\phi$ are learnable parameters. 
The policy is continuously improved alongside the model to jointly optimize the expected loss
$$\Lsample = \E_{w \sim \pi(\cdot | \phi)} [\Ltotal].$$
During training, the sampling policy is initialized to explore randomly at first, and converges to more promising parameter region over time.


Compared to DO, this formulation is more general in that
$\Lcost$ can be any differentiable cost function, and $\policy$ can be modeled as more sophisticated distribution to capture dependencies between variables.

Nevertheless, computing the gradients of $\Lsample$ analytically by enumerating all possible $w$ is generally intractable, depending on structure of $\policy$. Estimating the expectation from samples of $w$, on the other hand, does not provide gradients w.r.t $\phi$ for updating $\policy$.
Fortunately, using the identity $\nabla_{\phi} \policy = \policy \nabla_{\phi} \log \policy$, we can rewrite the gradient as
\begin{align}
\begin{split}
&\nabla \Lsample = \nabla \sum_{w} \policy{} \cdot \Ltotal \\
&= \sum_{w} \nabla \policy \cdot \Ltotal + \policy \cdot \nabla \Ltotal \\
&= \E_{w \sim \pi(\cdot | \phi)}[\nabla \log \policy \cdot \Ltotal + \nabla \Ltotal]\label{eq:policy-gradient}.
\end{split}
\end{align}

Notice that the final form is the sum of the original $\Ltotal$ gradient and a term involving gradient of the sampling distribution, and their expectation can be estimated from batch samples.
After training, we output the model that corresponds to maximum likelihood $w$.

\section{Experiments}
We run our experiments on different BERT models and tasks to evaluate our proposed methods. We initialize $w$ to be the baseline BERT network. 

To study the effect of model modifications, we also re-train the models with selected architecture for comparison. These models are given a ``-R" suffix in the experiments.
All of our models use bfloat16 and run inference on batches of 16 on  TPU-v2 hardware.

\subsection{Additional Input}
To use our model, we need two additional pieces of information. First, one needs to run profiling of a base model once to estimate the cost $c_i$ of each component. Note that the component costs can vary depending on the sequence lengths. From Table~\ref{tab:bert_base_cost}, we observed that the costs are similar across different sequence lengths, so we just use the highest cost across all profiled sequence length. Second, we need to decide on an acceptable metric drop. Once we do, we can increase $\lambda$ until the drop becomes unacceptable.

\subsection{Hyperparameters}

All hyperparameters, including number of training steps, are the same for the selected models and the base models for fair comparison. See Appendix~\ref{appendix:training} for more details.


\subsection{BERT Base}
We consider training the BERT-base structure in two different scenarios: English-only BERT and multilingual BERT, and investigate the effectiveness of the methods proposed above under different settings.

Regarding search space, we divide each component into two equally sized parts. We measure TPU run time of each component as shown in Table~\ref{tab:bert_base_cost} to compute the $c_i$'s above. We report the inference time averaged across sequence lengths to report speedup.

\begin{table}[t]
\small
    \centering
    \begin{tabular}{lrrr}
\toprule
               & \multicolumn{3}{c}{Sequence Length} \\
Component &  32 & 128 & 512 
\\\midrule
Feedforward  & 43.3\%& 58.6\%& 51.0\% \\
Attention Head &  54.9\%& 40.6\%& 48.7\% \\
Query-Key Similarity & 28.9\%& 20.6\%& 21.6\% \\
Attention Value & 22.8\%& 19.9\%& 21.6\% \\
Layer Normalization Mean &  0.8\% & 0.8\% & 0.7\% \\
Vertical Feedforward &  0.9\% & 1.3\% & 0.1\% 
\\\bottomrule
\end{tabular}
\caption{The computation time of each category of components in a full BERT-base network.}
\label{tab:bert_base_cost}
\end{table}

\subsubsection{English BERT}
We follow \citet{DBLP:journals/corr/abs-1810-04805}'s setting to pre-train English BERT-base.  We pick the fastest model with MNLI  dev set accuracy drop less than 1\%.

We evaluate our model on three datasets of the GLUE benchmark \cite{wang2019glue}. Table~\ref{tab:enbert_results} shows that SDO-R is 1.31 times faster with comparable quality to the baseline model.  

\begin{table}[t]
\small
    \centering
    \begin{tabular}{l|rrr|r}
\toprule
& \multicolumn{3}{c}{Task Metric} &\\
Model & MNLI & MRPC & SST2 & Speed 
\\\midrule
BERTBase & {\bf 84.5}\% &  83.0\% & {\bf 93.7}\% & 1  \\
DO       & 83.2\% & 82.3\% & 93.4\% & 1.06\\
DO-R     & {\bf 84.5}\% & 83.1\% & 93.6\% & 1.06 \\
SDO      & 83.0\% & {\bf 83.5}\% & 92.8\% & {\bf 1.31}\\
SDO-R    & 84.0\% & 82.4\% & 93.5\% & {\bf 1.31}
\\\bottomrule
\end{tabular}
\caption{English benchmark tasks and performance metrics of the base model and the selected models.}
\label{tab:enbert_results}
\end{table}


\subsubsection{Multilingual BERT}
We pre-train mutilingual BERT-base using SentencePiece \cite{DBLP:journals/corr/abs-1808-06226} and 120k vocab size on Wikipedia.  We pick the fastest model with pretraining  dev set accuracy drop less than 1\%.
We evaluate our models on two datasets of the XTREME benchmark \cite{Hu2020} for zero-shot learning. Table~\ref{tab:mbert_results} show that we can get a 14\% faster model with similar accuracy to the baseline model after retraining.

\subsubsection{Architecture Choices}

\begin{figure}[bt!]
 \centering
\includegraphics[width=3.5cm]{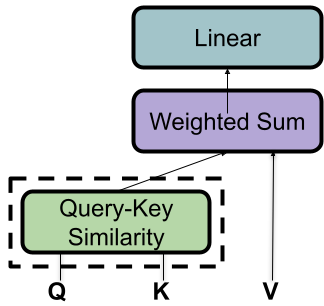}
\caption{Value Mean Pooling: instead of doing standard self-attention, we can remove query-key similarity in the dashed line. With that, we get a uniformly weighted average of values across the sequence.}
\label{fig:value_mean_pooling}
\end{figure}

\begin{figure*}[hbt!]
 \centering
\includegraphics[width=16cm]{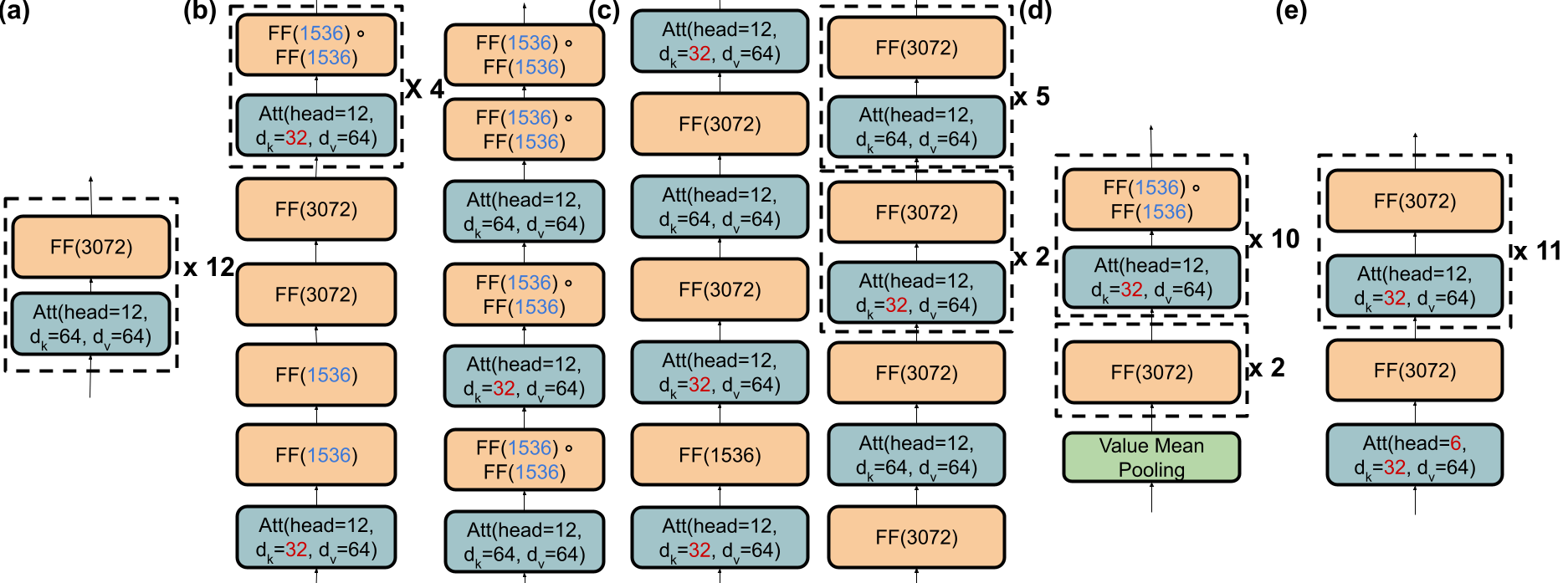}
\caption{Selected architectures: (a) BERT-base. (b) English BERT selected by SDO. (c) English BERT selected by DO. (d) multilingual BERT selected by SDO. (e) multilingual BERT selected by DO. For illustration simplicity, we omit the layernorm after each block in this figure. {\bf All} of the zero-mean normalization in layernorms are removed.}
\label{fig:bert_architecture}
\end{figure*}

The selected architectures are shown in Figure~\ref{fig:bert_architecture}. We can see English BERT and multilingual BERT have different network architectures chosen. Across all selected models and different pre-training tasks, some observations are
\begin{itemize}
\setlength\itemsep{0.001em}
    \item Zero-mean layer normalization was never chosen, raising doubts about its effectiveness. 
    \item Sometimes, the whole query-key similarity branch of an attention can be dropped, especially in earlier layers, making it a \emph{Value Mean Pooling} component as shown in Figure~\ref{fig:value_mean_pooling}.
    \item Attention and feedforward components can be dropped at the bottom and the top of the model.
    \item $\dk$ can be smaller, but not $\dv$.
    \item Vertically connected feedforward is better.
\end{itemize}

\begin{table}[t]
\small
    \centering
    \begin{tabular}{l|rr|rr}
\toprule
& \multicolumn{2}{c}{Task Metric} & \multicolumn{2}{c}{Performance Metric} \\
Model & XNLI & WikiAnn & Speed & \# Params \\\midrule
BERTBase & 70.3\% & 68.7\% & 1 & 172M \\
DO & 69.3\% & 65.5\%& 1.08 & 165M \\
DO-R & \bf{71.4\%}& \bf{70.4\%}& 1.08 & 165M \\
SDO & 70.3\% & 65.7\%& \bf{1.14} & \bf{161M} \\
SDO-R & 70.2\% & 69.6\% & \bf{1.14} & \bf{161M}
\\\bottomrule
\end{tabular}
\caption{Multilingual tasks metrics and performance metrics of the base model and the selected models.}
\label{tab:mbert_results}
\end{table}
 
\subsection{MiniBERT}
Distillation is a very effective approach to reduce model size and increase model speed if one can access a large amount of unlabeled data.
However, finding an efficient and accurate distilled model architecture can be difficult and may require exhaustive search. Here, we apply our model architecture search method to shrink a previous state-of-the-art distilled model for part-of-speech tagging and morphology  \cite{tsai-etal-2019-small} and show that we can make the model more efficient with our one-shot search algorithms.

\subsubsection{Model Profile and Search Space}

We profiled MiniBERT with results in Table~\ref{tab:minibert_cost}. Notice that, unlike BERT-base, the operations that are expensive here are different: vertical feedforward connection and layernorm are relatively more expensive.

\begin{table}[t]
\small
    \centering
    \begin{tabular}{lrrr}
\toprule
               & \multicolumn{3}{c}{Sequence Length} \\
Component &  32 & 128 & 512 
\\\midrule

Feedforward  & 32.2\%	&36.2\%	&30.2\% \\
Attention Head &  41.2\%&36.7\%	&47.1\% \\
Query-Key Similarity & 21.3\%&16.4\%&21.3\% \\
Attention Value &  18.9\%&15.5\%&21.3\% \\
Layer Normalization Mean &   6.6\%&6.4\%&4.6\% \\
Vertical Feedforward & 19.1\%&22.4\%&14.7\%
\\\bottomrule
\end{tabular}
\caption{The computation time of each category of components in a full MiniBERT network.}
\label{tab:minibert_cost}
\end{table}

Regarding the search space, we divide the feedforward layer to eight equally-sized components. Each query-key similarity and attention value are divided into two components. Each attention head is one component.

\begin{table}[t]
\small
    \centering
    \begin{tabular}{l|r|rr}
\toprule
Model & Accuracy  & Speed & \# params
\\\midrule
Teacher  & 94.3\% & 1 & 172M
\\\midrule
\citet{tsai-etal-2019-small}  & 93.7\% & 20 & 33M  \\
MiniBERT (Ours) & \bf{94.1\%} & 20 & 33M
\\\midrule
DO  & 93.5\% & 34& 32M  \\
DO-R  & 93.7\% & 34&  32M \\
SDO  & 93.7\% & \bf{36}  & 31M\\
SDO-R  & \bf{93.8}\% & \bf{36}  & \bf{31M}
\\\bottomrule
\end{tabular}
\caption{Multilingual part-of-speech tagging accuracy and performance metrics of the base model and the selected models.}
\label{tab:mini_bert_pos_results}
\end{table}

\begin{table}[t]
\small
    \centering
    \begin{tabular}{l|r|rr}
\toprule
Model & Accuracy  & Speed & \# params
\\\midrule
Teacher  & 91.1\% & 1 & 172M
\\\midrule
\citet{tsai-etal-2019-small}  & 88.6\% & 20 & 33M  \\
MiniBERT (Ours) & \bf{90.7\%} & 20 & 33M
\\\midrule
DO  & 89.8\% & 26 &  33M \\
DO-R  & \bf{90.4}\% & 26 & 33M  \\
SDO  & 90.2\% & \bf{33}&  32M \\
SDO-R  & 90.2\% & \bf{33} & 32M
\\\bottomrule
\end{tabular}
\caption{Multilingual morphology accuracy and performance metrics of the base model and the selected models.}
\label{tab:mini_bert_morph_results}
\end{table}

\subsubsection{More Accurate Distilled Model}
First, we found the distilled model trained by \cite{tsai-etal-2019-small} can be improved by better distillation techniques:  we fix the teacher model and improve the distilled model by removing all dropouts and apply linear weight ramp-up of labeled data during distillation, closing more than half of the distillation gap.

\begin{figure*}[hbt!]
 \centering
\includegraphics[width=13cm]{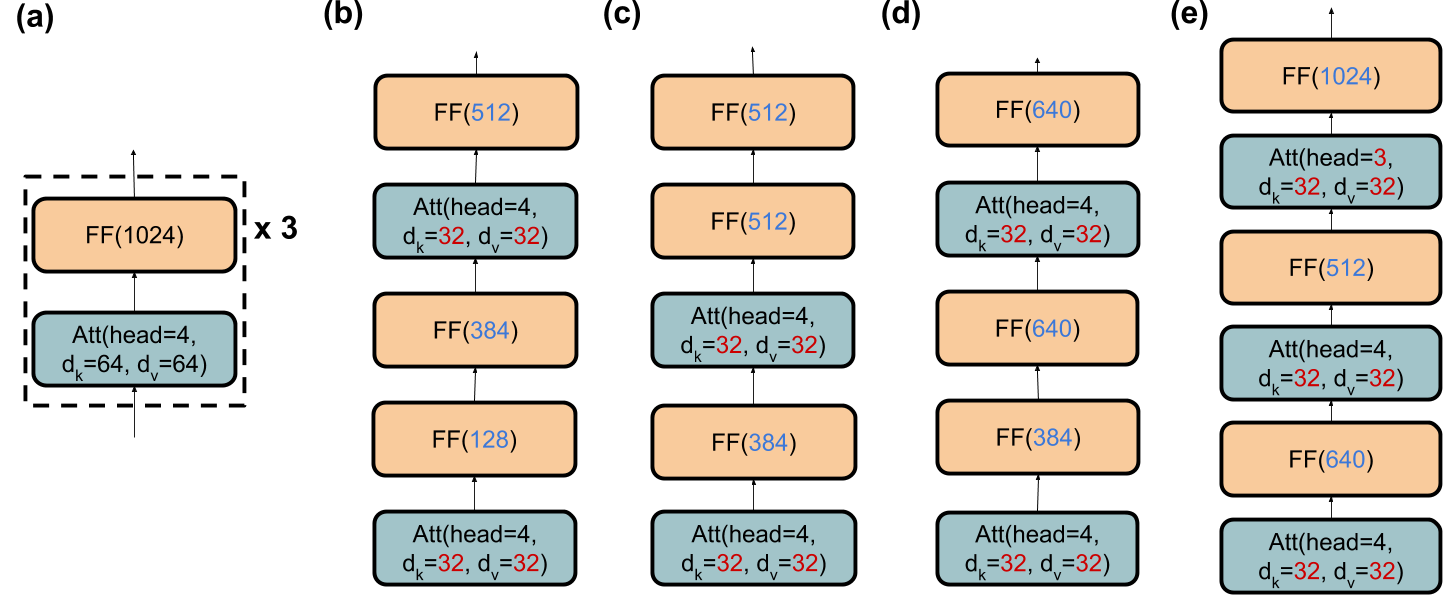}
\caption{Distilled BERT architectures comparison: (a) MiniBERT architecture (b) Selected architecture by SDO for the part-of-speed task. (c) Selected architecture by DO for the part-of-speed task. (d) Selected architecture by SDO for the morphology task. (e) Selected architecture by DO for the morphology task. For illustration simplicity, we omit the layernorm after each block in this figure. {\bf All} of the zero-mean normalization in layernorms are removed.}
\label{fig:distilled_architecture}
\end{figure*}

\subsubsection{Even Faster Distilled Model}
We search for the fastest distilled model with dev  accuracy drop less than 0.3\%. Table~\ref{tab:mini_bert_pos_results} and Table~\ref{tab:mini_bert_morph_results} show that the MiniBERT is already 20 times faster than the BERT-base. We show our selected models can further improve the distilled model to be 1.7 times faster than the state-of-the-art distilled model and 33 to 36 times faster than the base model with small accuracy drop. The scale of the change is much larger than pretrained models where we see about 1.1 to 1.2 times speedup.  We conjecture this is due to that BERT needs a lot of model capacity to learn the pre-training tasks and it is difficult to achieve bigger speedups there. 

All the selected architecture are shown in Figure \ref{fig:distilled_architecture}. We observed that different architectures are selected for different tasks. The morphology task, which has 1000 times more classes than part-of-speech tagging, needs a bigger model to keep the high accuracy.

\subsection{Comparing One-Shot Search Algorithms}
Observing the results above, we can see that SDO, which optimizes the speed objective directly without relaxation, usually achieve bigger speedup than DO given the same model quality constraint.

In the pretraining cases, re-training may be needed depending on the downstream task. We conjecture this is because needing to make architecture exploration makes the models not have enough effective training steps as reported by previous work \cite{DBLP:journals/corr/abs-1907-11692} to achieve SOTA accuracy. In the distillation case, we see that SDO has the same quality as SDO-R, so we can remove the retraining step and save 2 times the resources for training another model.  

\section{Conclusion}

We have described a way to define the model architecture space of the Transformer based on component composition, and we have proposed a sample-based one-shot search algorithm to find efficient model architectures efficiently. We show empirically that our methods work well with both BERT-base and an already-small distilled BERT on a variety of tasks.


\bibliographystyle{acl_natbib}
\bibliography{main}

\newpage
\appendix

\section{Training Setting}
\label{appendix:training}
\subsection{Implementation Formulation}
In our SDO implementation, we model the sampling policy as Bernoulli distribution with mean parameter represented by $\E[w_i] = \textrm{Sigmoid}(\phi_i)$ and that
$$\log \policy[w_i] \propto (-1)^{1-w_i}\phi_i.$$

To allow different update size to $\policy$ relative to $\Ltotal$, we modified Equation~\ref{eq:policy-gradient} to introduce a tuneable hyperparameter $\nu$. The final gradients computed on sampled batch data is given by
\begin{align*}
\nu (-1)^{1-w_i} \nabla \phi_i \cdot \Ltotal + \nabla \Ltotal,
\end{align*}
where
\begin{align*}
\Ltotal = \Lorig + \lambda \sum_i w_i c_i.
\end{align*}


\subsection{One-Shot Search Algorithm Tuning}
We do a grid search with $\nu = \{0.001, 0.01\}$ and $\lambda = \{0.01, 0.001, 0.0001, 0.00001, 0.000001\}$ to find the fastest model with an acceptable accuracy.

We  do a comprehensive search for English BERT to understand the hyperparameter impacts better. The results are in Table~\ref{tab:mnli_dev_all_models}. Overall, we see that as we increase $\lambda$, the quality start to gradually decrease and some components are dropped. Then, after a certain point, the model will collapse with really low accuracy and most of the components dropped for both DO and SDO. 

From our experiments, we also found that $\lambda = 0.001$ for DO, and $(\lambda, \nu) = (0.00001, 0.01)$ for SDO work out-of-the-box for other experiments. We conjecture this is because the model is relative stable when the hyperparameters are at the "saddle" area. Thus, in most of our experiments, we use the above hyperparameter values, and only retune if the model quality is off. 

\subsection{Pretraining Details}
For English BERT, we use the following pretraining hyperparameters:
\begin{itemize}
    \item Pretraining steps: 250k (90\% sequence length 128, then 10\% sequence length 512.)
    \item Public BERT wordpieces.
    \item Batch size: 4096
    \item Optimizer: LAMB \cite{You2020Large}
    \item Learning rate: 0.0018
    \item Num warmup steps: 2500
\end{itemize}
For multilingual-BERT, we use the following hyperparameters:
\begin{itemize}
    \item Sequence length: 128
    \item Num vocabs: 120k
    \item Tokenization: sentecnepiece
    \item Pretraining steps: 1M
    \item Batch size: 4096
    \item Optimizer: LAMB \cite{You2020Large}
    \item Learning rate: 0.0018
    \item Num warmup steps: 1250
\end{itemize}

We notice that while multilingual BERT results match the state-of-the-art but the English BERT does not. This may be fixed by training longer, but it should not affect our neural architecture search study.

\subsection{Distillation Details}
Compared to \cite{tsai-etal-2019-small}, we made a couple modifications to the distillation algorithm. First, we do not use logits in distillation. We find using logits does not help  model quality but make the distillation pipeline run much longer due to passing huge logits tensors. Thus, we remove the logits loss computation and just use the silver labels generated by the teacher model. That enables us to train the models for longer in less time. Second, we remove all the dropouts in the model to make the student have more model capacity and overfit the teacher better. Finally, we linearly rampup the weight ratio of the labeled data from zero after the training has progressed $p\%$ and stop at $q\%$ (meaning we only use labeled data after that). The intuition is to let the student model slowly adapt to the gold data distribution. The idea is similar to Teacher Annealing \cite{DBLP:journals/corr/abs-1907-04829}.

To summarize, here are the hyperparameters
\begin{itemize}
    \item Distillation data: de-duplicated multilingual Wikipedia without upsampling.
    \item Train steps: 2M
    \item Learning rate: 0.0005
    \item Optimizer: ADAM
    \item Warmup steps: 10k
    \item Batch size: 768 (704 silver and 64 gold in each batch.)
    \item p = 80, q = 100
\end{itemize}


\section{Derivations for Connecting Residual Components}
\label{appendix:res_to_conn}
Here we show the equivalence between $\Res(\FF_d)$ and $m$ horizontally connected $\FF_{\frac{d}{m}}$.
The equivalence between $\Res(\Att{a,\dk,\dv})$ and $a$ horizontally connected $\Att{1,\dk,\dv}$ follows the same logic.

One horizontal connector puts $\FF_{\frac{d}{m}}$ at the second output:
\begin{align*}
&\Conn\left(\FF_{\frac{d}{m}}, 0\right)(X, 0) = \left(X, \FF_{\frac{d}{m}}(X)\right)
\end{align*}
When we horizontally connect another component, both $\FF_{\frac{d}{m}}$ are accumulated at the second output.
\begin{align*}
&\left(\Conn\left(\FF_{\frac{d}{m}}, 0\right) \circ\Conn\left(\FF_{\frac{d}{m}}, 0\right)\right)(X, 0) \\
&= \left(X, \FF_{\frac{d}{m}}(X)+\FF_{\frac{d}{m}}(X)\right)
\end{align*}
By repeating $m$ times, the second output becomes $\FF_d(X)$.
\begin{align*}
&\left(\mathop{\bigcirc}\limits_{i=1}^{m}\Conn(\FF_{\frac{d}{m}}, 0)\right)(X, 0) \\
&= \left(X, \sum_{i=1}^m \FF_{\frac{d}{m}}(X)\right)  = (X, \FF_d(X))
\end{align*}
Finally, we can combine both output using $\Combine(X,R)=X+R$, and the equivalence is straightforward.
\begin{align*}
\begin{split}
\Res(\FF_d)(X) &= X + \FF_d(X) \\
&=\Combine(X, \FF_d(X)) \\
&=\left(\Combine\circ\left(\mathop{\bigcirc}\limits_{i=1}^{m}\Conn(\FF_{\frac{d}{m}}, 0)\right)\right)(X)
\end{split}
\end{align*}

\section{Example of Connector Units}
\label{appendix:connector-example}
\begin{figure*}[bt]
 \centering
\includegraphics[width=12cm]{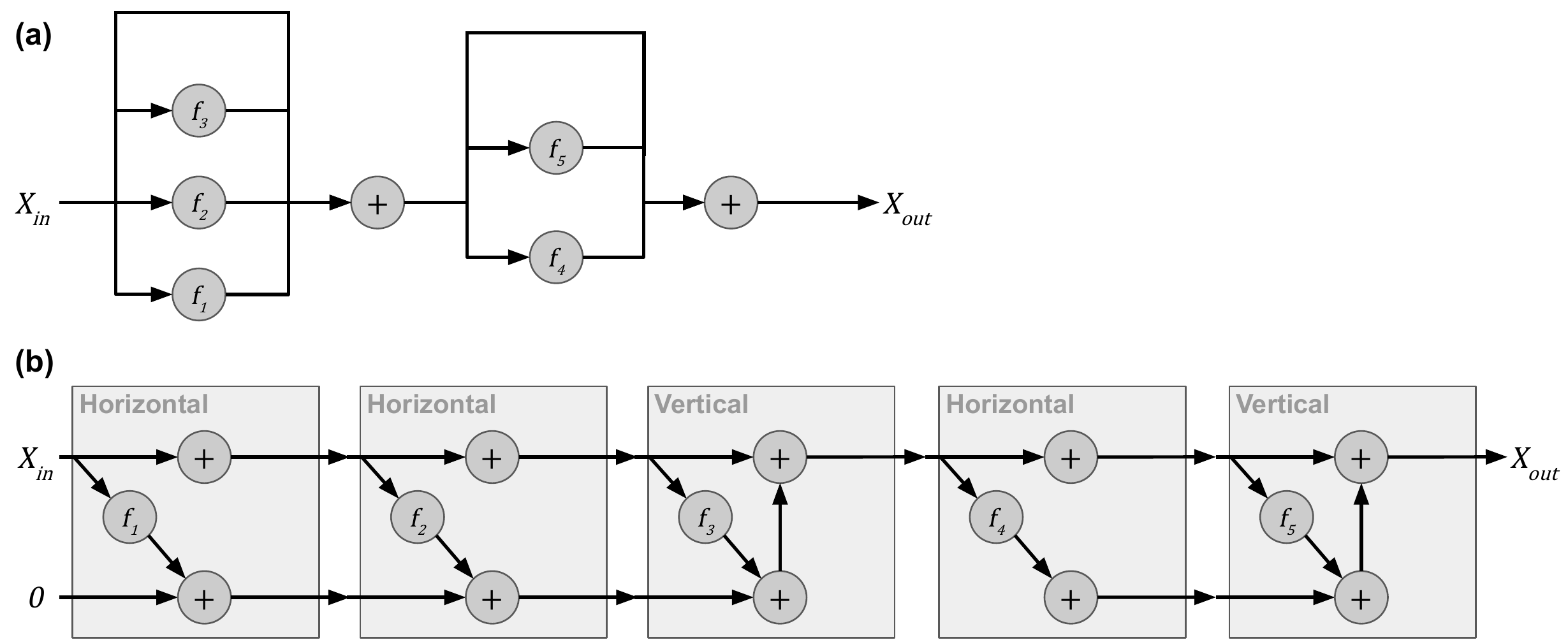}
\caption{(a) An example $2$-layer network. (b) Equivalent network expressed in connector units.}
\label{fig:connector-example}
\end{figure*}

Figure~\ref{fig:connector-example} shows a $[3,2]$ residual network and its equivalent expression in connector units $(\Conn(f_5, 1)\circ\Conn(f_4, 0)\circ\Conn(f_3, 1)\circ\Conn(f_2, 0)\circ\Conn(f_1, 0))(X, 0)$.

\section{Detailed Experiment Results}
Here, we provide detailed experiment results and the final model hyperparameters.

\subsection{English BERT}
We report the hyperparameters and the MNLI dev set accuracy used to select the best model architecture of all tasks in Table~\ref{tab:mnli_dev_all_models}. After getting the best model architecture, we search over learning rate \{2e-5, 3e-5, 4e-5\}, train epochs \{6, 7\} on GLUE data sets to find the model with the best dev set accuracy. We report the test set accuracy of the best models in Table~\ref{tab:enbert_results_dev_test}.

\begin{table}
\small
\centering
\begin{tabular}{lllrr}
\toprule
Algorithm & $\lambda$ &  $\nu$ & Accuracy & Speed
\\\midrule
BertBASE & & & 82.8\%& 1\\
DO  & 1e-2 & & 79.1\% & 1.3\\
{\bf DO} & 1e-3 & & 82.2\%& 1.06\\
DO  & 1e-4 & & 82.6\%& 1.03\\
DO  & 1e-5 & & 83.1\%& 1.03\\
SDO  & 1e-4 & 1e-2 & 74.4\%& 3\\
{\bf SDO}  & 1e-5 & 1e-2 & 82.2\%& 1.31\\
SDO  & 1e-6 & 1e-2 & 83.1\%& 1
\\\bottomrule
\end{tabular}
\caption{MNLI dev set accuracy of different model architectures. The models are trained for 125k steps. The selected model is marked bold.}
\label{tab:mnli_dev_all_models}
\end{table}

\begin{table*}[t]
\small
    \centering
    \begin{tabular}{l|rrr|rr}
\toprule
& \multicolumn{3}{c}{Task Metric} & \multicolumn{2}{c}{Performance Metric}\\
Model & MNLI & MRPC & SST2 & Speed & \# Params
\\\midrule
BERTBase & 84.4\% / {\bf 84.5}\% &  86.3\% / 83.0\% & 91.4\% / {\bf 93.7}\% & 1  & 110M \\
DO       & 83.2\% / 83.2\% & 86.8\% / 82.3\% & 91.6\% / 93.4\% & 1.06 & 106M\\
DO-R     & 84.8\% / {\bf 84.5}\% & 85.7\% / 83.1\% & 92.1\% / 93.6\% & 1.06 & 106M\\
SDO      & 83.4\% / 83.0\% & 87.3\% / {\bf 83.5}\% & 91.6\% / 92.8\% & {\bf 1.31} & 98M\\
SDO-R    & 83.6\% / 84.0\% & 86.0\% / 82.4\% & 92.4\% / 93.5\% & {\bf 1.31} & 98M
\\\bottomrule
\end{tabular}
\caption{English BERT: detailed (dev/test) benchmark tasks metrics and performance metrics of the base model and the selected models.}
\label{tab:enbert_results_dev_test}
\end{table*}

\subsection{Multilingual BERT}
The fatest DO model with acceptable accuracy drop uses $\lambda = 0.01$ at 60\% MLM accuracy. The best SDO model uses $\lambda = 10^{-5}$ and $\nu = 0.01$ with 60\% MLM accuracy. The BERT base model without model selection has 61\% MLM accuracy.

We compiled detailed fine-tuning results in Table~\ref{tab:mbert_results_dev_test}.

\begin{table*}[t]
\small
    \centering
    \begin{tabular}{l|rr|rr|rr}
\toprule
& \multicolumn{2}{c}{Task Metric} & \multicolumn{2}{c}{Performance Metric} & \multicolumn{2}{c}{XNLI Hyperparameters}\\
Model & XNLI Accuracy & WikiAnn F1 & Speed & \# Params &  epochs &  learning rate \\\midrule
BERTBase & 70.7\% / 70.3\% & 68.3\% / 68.7\% & 1 & 172M & 3 & 3e-5\\
DO & 69.0\% / 69.3\% & 65.1\% / 65.5\%& 1.08 & 165M& 3 & 5e-5\\
DO-R & 71.2\% / \bf{71.4\%}& 70.2\% / \bf{70.4\%}& 1.08 & 165M & 3 & 2e-5\\
SDO & 70.3\% / 70.3\% & 65.5\% / 65.7\%& \bf{1.14} & \bf{161M} & 3 & 3e-5\\
SDO-R & 70.3\% / 70.2\% & 69.3\% / 69.6\% & \bf{1.14} & \bf{161M} & 3 & 3e-5
\\\bottomrule
\end{tabular}
\caption{Multilingual BERT: Detailed (dev/test) benchmark tasks metrics and performance metrics of the base model and the selected models. For WikiAnn, all runs train for 10 epochs with learning rate 3e-5.}
\label{tab:mbert_results_dev_test}
\end{table*}




\end{document}